\pdfoutput=1
\documentclass{article}
    \PassOptionsToPackage{numbers, compress}{natbib}

    \usepackage[final]{neurips_2020}

\usepackage[utf8]{inputenc} 
\usepackage[T1]{fontenc}    
\usepackage{hyperref}       
\usepackage{url}            
\usepackage{booktabs}       
\usepackage{amsfonts}       
\usepackage{nicefrac}       
\usepackage{microtype}      
\usepackage{amsmath}

\newcommand{\T}[1]{{\mathcal{#1}}} 
\newcommand{\V}[1]{{\mathbf{#1}}} 
\usepackage{graphicx}
\usepackage{enumitem}
\usepackage{comment}
\usepackage{enumitem}
\usepackage{wrapfig}
\usepackage{xcolor}
\usepackage{subfig}

\usepackage{subfloat}
\newcommand{\acedit}[1]{{\color{teal} #1}}
\newcommand{\ac}[1]{\acedit{[AC: #1]}}

\newcommand{\ours}{\textsc{dive}}

\renewcommand{\ac}[1]{}
\title{Learning Disentangled Representations of   \\ Video  with Missing Data}

\author{%
   Armand Comas Massagué$^1$ \\
  \texttt{comasmassague.a@northeastern.edu}
  \And
  Chi Zhang$^2$ \\
  \texttt{zhang.chi13@northeastern.edu}
  \And
  Zlatan Feric$^2$  \\
  \texttt{feric.z@northeastern.edu}
  \And
  Octavia Camps$^1$ \\
  \texttt{camps@coe.neu.edu}
  \And
  Rose Yu$^{2,3}$  \\
  \texttt{roseyu@ucsd.edu} \thanks{$^1$College of Electrical and Computer Engineering, $^2$ Khoury College of Computer Sciences, Northeastern University, MA, USA, $^3$Computer Science \& Engineering, University of California San Diego, CA, USA.}
}

\begin{document}
\maketitle

\begin{abstract}
Missing data poses significant challenges while learning representations of video sequences. We present Disentangled Imputed Video autoEncoder (\textsc{dive}), a deep generative model that imputes and predicts future video frames in the presence of missing data. Specifically, \ours{} introduces a missingness latent variable,  disentangles the hidden video representations into static and dynamic appearance, pose, and missingness factors for each object.  \ours{} imputes each object's trajectory where the data is missing. 
%
On a moving MNIST  dataset with various missing scenarios,  \ours{} outperforms the state of the art baselines by a substantial margin.  We also present comparisons on a real-world MOTSChallenge pedestrian dataset, which demonstrates the practical value of our method in a more realistic setting.  Our code and data can be found at \url{https://github.com/Rose-STL-Lab/DIVE}.
\end{abstract}

\section{Introduction}

Videos contain rich structured  information about our physical world. Learning representations from video enables intelligent machines to reason about the surroundings and it is essential to a range of tasks in machine learning and computer vision, including  activity recognition \cite{karpathy2014large},  video prediction  \cite{mathieu2016deep} and spatiotemporal reasoning \cite{jang2017tgif}.   One of the fundamental challenges in video representation learning is the high-dimensional,  dynamic, multi-modal  distribution of pixels. 
Recent research in deep generative models  \cite{denton2017unsupervised,ddpae, kosiorek2018sequential, ye2019compositional} tackles the challenge by exploiting inductive biases of videos and projecting the high-dimensional data into substantially lower dimensional space. These methods search for \textit{disentangled} representations by decomposing the latent representation of video frames into semantically meaningful factors \cite{locatello2018challenging}.


Unfortunately, existing methods cannot reason about the objects when they are missing in videos. In contrast, a five month-old child can   understand that objects continue to exist even when they are unseen, a phenomena known as ``object permanence'' \cite{baillargeon1985object}. Towards making intelligent machines, we study learning disentangled representations of videos with missing data. We consider a variety of missing scenarios that might occur in natural videos: objects can be partially occluded; objects can disappear in a scene and reappear; objects can also become missing while changing their size, shape, color and brightness. The ability to disentangle these factors and learn appropriate representations is an important step toward spatiotemporal decision making in complex environments.

In this work, we build on the deep generative model of \textsc{ddpae} \cite{ddpae} which integrates structured graphical models into deep neural networks. Our model, which we call Disentangled-Imputed-Video-autoEncoder (\textsc{dive}),  (\textbf{i}) learns representations that factorize into appearance, pose and missingness latent variables; (\textbf{ii}) imputes missing data by sampling from the learned latent variables; and (\textbf{iii}) performs unsupervised stochastic video prediction using the imputed hidden representation. Besides imputation, another salient feature of our model is (\textbf{iv}) its ability to robustly  generate objects even when their appearances are changing  by modeling the static and dynamic appearances separately.  Thismakes our technique more applicable to  real-world problems.


We demonstrate the effectiveness of our method on a moving MNIST dataset with  a variety of missing data scenarios including partial occlusions, out of scene, and missing frames with varying appearances. We further evaluate on the Multi-Object Tracking and Segmentation (MOTSChallenge)  object tracking and segmentation challenge dataset. We show that \textsc{dive} is able to accurately infer missing data, perform video imputation and reconstruct input frames and generate future predictions. Compared with baselines, our approach is robust to  missing data and achieves significant improvements in video prediction performances.

\section{Related Work}


\paragraph{Disentangled Representation.}
Unsupervised learning of disentangled representation for sequences generally falls into three categories: VAE-based 
\cite{hsu2017unsupervised,kosiorek2018sequential, ddpae, ye2019compositional, kossen2020structured, stanic2019r},   GAN-like models \cite{villegas2017decomposing, denton2018stochastic, denton2017unsupervised, kurutach2018learning} and Sum-Product networks \cite{kossen2020structured, pmlr-v97-stelzner19a}. For video data, a common practice is to encode a video frame into latent variables and  disentangle the latent representation into \textit{content} and \textit{dynamics} factors. For example, \cite{ddpae}  assumes the content (objects, background) of a video is  fixed across frames, while the position of the content can change over time. In most cases, models can only handle complete video sequences without missing data.
One  exception is  \textsc{sqair} \cite{kosiorek2018sequential}, an generalization of  \textsc{air} \cite{Eslami2016AttendIR}, which makes use of a latent variable to explicitly encode the \textit{presence} of the respective object. \textsc{sqair} is further extended to an accelerated  training scheme \cite{pmlr-v97-stelzner19a} or to better encode relational inductive biases \cite{kossen2020structured, stanic2019r}.
However, \textsc{sqair} and its extensions have no mechanism to recall an object. This leads to discovering an object as new when it reappears in the scene. 

\paragraph{Video Prediction.}
Conditioning on the past frames, video prediction models are trained to reconstruct the input  sequence and  predict future frames.  Many video prediction methods use dynamical modeling \cite{liu2018dyan} or deep neural networks to learn a deterministic transformation from input to output, including LSTM \cite{srivastava2015unsupervised},  Convolutional LSTM \cite{finn2016unsupervised} and PredRNN \cite{wang2018predrnn++}. These methods often suffer from blurry predictions and cannot properly model the inherently uncertain future \cite{kalchbrenner2017video}.
In contrast to deterministic prediction, we prefer stochastic video prediction  \cite{mathieu2016deep, xue2016visual, kalchbrenner2017video, babaeizadeh2018stochastic, denton2018stochastic, wangprobabilistic}, which is more suitable for capturing the stochastic dynamics of the environment. For instance, \cite{kalchbrenner2017video} proposes an auto-regressive model to generate pixels sequentially.   \cite{denton2018stochastic} generalizes \textsc{vae} to video data with a learned prior. \cite{kumar2019videoflow} develops a normalizing flow video prediction model. \cite{wangprobabilistic} proposes a Bayesian Predictive Network to learn the prior distribution from noisy videos but without disentangled representations. 
Our main goal is to learn  disentangled latent representations from video that are both interpretable and robust to missing data.   


\paragraph{Missing Value Imputation.}
Missing value imputation is the process of replacing the missing data in a sequence by an estimate of its true missing value. It is a central challenge of sequence modeling. Statistical methods often impose strong assumptions on the missing patterns. For example,  mean/median averaging  \cite{acuna2004treatment} and  MICE \cite{buuren2010mice}, can only handle data missing at random.  Latent variables models with the EM algorithm \cite{nelwamondo2007missing} can impute data missing not-at-random but are restricted to certain parametric models.  Deep generative models offer a flexible framework of missing data imputation. For instance, \cite{yoon2018estimating, che2018recurrent, cao2018brits} develop variants of recurrent neural networks to impute time series.   \cite{yoon2018gain,luo2018multivariate,liu2019naomi} propose GAN-like models to learn  missing patterns in multivariate time series.
Unfortunately, to the best of our knowledge, all recent developments in generative modeling for missing value imputation have focused on low-dimensional time series, which are not directly applicable to high-dimensional video with complex scene dynamics.

\section{Disentangled-Imputed-Video-autoEncoder (DIVE)}



\begin{figure}[t]
    \centering
    \includegraphics[trim=60 360 60 20,clip,width=0.95\textwidth]{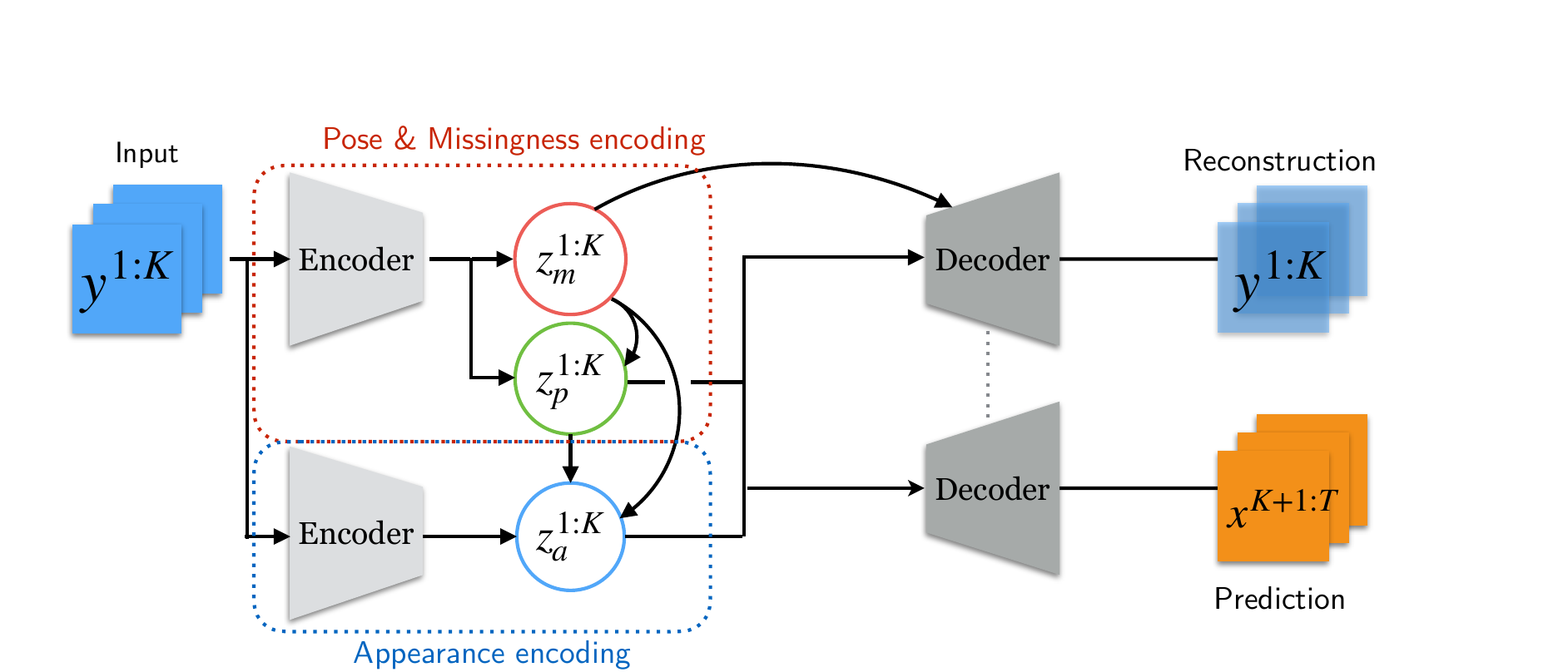}
    \caption{Overall architecture of \textsc{dive}, which takes the  input video with missing data, infers the missingness (red),  pose (green) and appearance (blue) latent variables. Two separate decoders reconstruct and predict the future sequences. The model is trained following the VAE framework.
    }\label{fig:arch_all}
    \vspace{-3mm}
\end{figure}

Videos often capture multiple objects moving with complex dynamics. %
For this work, we assume that each video has a maximum number of $N$ objects, we observe a video sequence up to $K$  time steps and aim to predict $T-K+1$ time steps ahead. 
The key component of \textsc{dive} is based on the decomposition and disentangling of the objects representations within a VAE framework, with similar recursive modules as in \cite{ddpae}.  Specifically, we decompose the objects in a video and  assign three sets of latent variables to each object: appearance, pose and missingness, representing distinct attributes. During inference, \textsc{dive} encodes the input video into latent representations,  performs sequence imputation in the latent space and updates the hidden representations. The generation model then samples from the latent variables to reconstruct and generate future predictions.  Figure \ref{fig:arch_all} depicts the overall  pipeline of our model. 

Denote a  video sequence with missing data as $(\V{y}^1,\cdots, \V{y}^t)$ where each $\V{y}^t \in \mathbb{R}^d$ is a frame. We assume an  object in a video consists of \textit{appearance}, \textit{pose} (position and scale), and \textit{missingness}. For each  object  $i$ in frame $t$, 
we aim to learn the latent representation $\V{z}_i^t$ and disentangle it  into three  latent variables:
\begin{align}
    \V{z}_i^t = [\V{z}_{i,a}^t, \V{z}_{i,p}^t, \V{z}_{i,m}^t], \quad \V{z}_{i,a}^t \in \mathbb{R}^h, \V{z}^t_{i,p}\in \mathbb{R}^3,\V{z}^t_{i,m} \in \mathbb{Z}
\end{align}
where  
$\V{z}_{i,a}^t$ is the {\it appearance} vector with dimension $h$, $\V{z}_{i,p}^t$ is the {\it pose} vector with $x$, $y$ coordinates and scale and $\V{z}_{i,m}^t$ is the binary {\it missingness} label.  $\V{z}_{i,m}^t = 1$ if the object is occluded or missing.


\subsection{Imputation Model}
The imputation model  leverages the  missingness  variable $\V{z}_{i,m}^t$ to update the  hidden states. When there is no missing data, the encoded hidden state, given the input frame,  is  
$\V{h}_{i,y}^{t} = f_{\text{enc}}(\V{h}_{i,y}^{t-1}, \V{h}_{i,y}^{t+1}, [\V{y}^t, \V{h}_{i-1,y}^{t}])$, where we enforce separate representations for each object. We implement the encoding function $f_{\text{enc}}$ with a bidirectional LSTM to propagate the hidden state over time. However, in the presence of missing data, such hidden state is unreliable and needs imputation. 
Denote the imputed hidden state as $\hat{\V{h}}_{i,y}^{t}$ which will be discussed shortly. We update a latent space  vector $\V{u}_i^t$ to select the corresponding hidden state, given the sampled missingness variable:
%
%
\begin{eqnarray}
   \V{u}_i^t =\begin{cases}
     \hat{\V{h}}_{i,y}^{t} & \V{z}_{i,m}^{t} =1\\
         \gamma \V{h}_{i,y}^{t} +  (1-\gamma)\hat{\V{h}}_{i,y}^{t} & \V{z}_{i,m}^{t}=0
   \end{cases}, \quad  \gamma \sim \text{Bernoulli}(p)
  \label{eq:imp}
\end{eqnarray}
%
Note that we apply a mixture of   input hidden state  $\V{h}_{i,y}^{t}$  and  imputed hidden state $\hat{\V{h}}_{i,y}^{t}$  with  probability $p$. In our experiments, we found this mixed strategy to be helpful in mitigating covariate shift \cite{bickel2009discriminative}. 
It forces the model to learn the correct imputation with self-supervision, which is reminiscent of the scheduled sampling \cite{bengio2015scheduled} technique for sequence prediction.


The pose hidden states $\V{h}_{i,p}^t$ are obtained by propagating the updated latent representation through an LSTM network $ \V{h}_{i,p}^{t}= {\texttt{LSTM}}( \V{h}_{i,p}^{t-1}, \V{u}^{t}_i)$. 
For prediction 
we use an LSTM network, with only ${h}_{i,p}^{t-1}$ as input in time $t$. 
We obtain the imputed hidden state  by means of auto-regression. This is based on the assumption that a video sequence is locally stationary and the most recent history is predictive of the future.  Given the updated latent representation at time $t$,  the imputed hidden state at the next time step  is:
\begin{equation}
  \hat{\V{h}}_{i,y}^{t} = \texttt{FC}(\V{h}_{i,p}^{t-1})\label{eq:h-hat}
\end{equation}
where $\texttt{FC}(\cdot)$ is  a fully connected layer.  This approach is similar in spirit to the time series imputation method in \cite{cao2018brits}. However, instead of imputing in the observation space, we perform imputation in the space of latent representations. 

\subsection{Inference Model}

\paragraph{Missingness Inference.}
For the  missingness variable $\V{z}_{i,m}^t$, we also leverage the input encoding. We use a heaviside step function to make it binary:
\begin{eqnarray}
\V{z}_{i,m}^t = {H}(x),\quad x\sim \T{N}(\V{\mu}_{m}, \V{\sigma}_{m}^{2}), \quad
[\V{\mu}_{m}, \V{\sigma}_{m}^{2}] = \texttt{FC}(\V{h}_{i,y}^t),
\quad H(x)=  \begin{cases}1 & x\geq 0\\ 0 & x<0\end{cases}
\end{eqnarray}
where  $\sigma$ is the standard deviation of the noise, which is obtained from the hidden representation.


\paragraph{Pose Inference.} The pose variable (position and scale) encodes the spatiotemporal dynamics of the video.  We follow the  variational inference technique  for state-space representation of sequences   \cite{chung2015recurrent}.  That is, instead of directly inferring $\V{z}_{i,p}^{1:K}$ for $K$ input frames, we use a stochastic variable $\beta_i^t$ to reparameterize the state transition probability:
    \begin{eqnarray}
q(\V{z}^{1:T}_{i,p}|\V{y}^{1:K})  = \prod_{t=1}^K q(\V{z}_{i,p}^{t}|\V{z}_{i,p}^{1:t-1}),\quad \V{z}_{i,p}^{t} = f_{\text{tran}}(\V{z}_{i,p}^{t-1},  \beta_i^t), \quad \beta^t_i \sim 
\T{N}(\V{\mu}_{p}, \V{\sigma}_{p}^{2}) 
    \end{eqnarray}
where the state transition $f_{\text{tran}}$ is a deterministic mapping from the previous state to the next time step.   The  stochastic transition variable $\beta^t_i$ is sampled from a Gaussian distribution parameterized by a mean $\mu_p$ and variance $\sigma_p^2$ with $[\V{\mu}_{p}, \V{\sigma}_{p}^{2}] = \texttt{FC}(\V{h}_{i,p}^t)$.

\paragraph{Dynamic Appearance.} Another novel feature of our approach is its ability to robustly generate objects even when their appearances are changing across frames. $\V{z}_{i,a}^t$ is the time-varying appearance. In particular, we decompose  the appearance latent variable into  a static component  $\V{a}_{i,s}$ and a dynamic component  $\V{a}_{i,d}$ which we model separately.  The static component captures the inherent semantics of the object while the dynamic component models the nuanced variations in shape. 
%

%
For the static component, we follow the procedure in \cite{ddpae} to perform inverse affine spatial transformation $\mathcal{T}^{-1}(\cdot ; \cdot)$, given the pose of the object to  center  in the frame and rectify the images with a selected crop size. Future prediction is done in an autoregressive fashion: 
\begin{eqnarray}
\V{a}_{i,s} = \texttt{FC}(\V{h}_{i,a}^K), \quad  
\V{h}_{i,a}^{t+1} =\begin{cases} \texttt{LSTM}_1(\V{h}_{i,a}^{t},  \mathcal{T}^{-1}(\V{y}^t; \V{z}_{i,p}^t))& t<K\\
\texttt{LSTM}_2(\V{h}_{i,a}^{t})& K \le t < T\end{cases} \label{eq:stat-app}
    \end{eqnarray}
 %
%
Here the appearance hidden state $\V{h}_{i,a}^{t}$ is propagated through an LSTM, whose last output is used to infer the static appearance. 
Similar to poses,  we  use a state-space representation for the
 dynamic   component, but directly model the  difference in appearances, which helps stabilizing training:
\begin{eqnarray}
 \V{a}_{i,d}^{1}=\texttt{FC}([\V{a}_{i,s}, \mathcal{T}^{-1}(\V{y}^1; \V{z}_{i,p}^1)]),\quad \V{a}_{i,d}^{t+1} = \V{a}_{i,d}^{t}+ \delta_{i,d}^{t}, \quad \delta_{i,d}^{t} = \texttt{FC}([\V{h}_{i,a}^{t}, \V{a}_{i,s}])
\end{eqnarray}
The final appearance variable is sampled from a Gaussian distribution parametrized by the concatenation of static and dynamic components, which are randomly mixed with a probability $p$: 
\begin{eqnarray}
 q(\V{z}_{i,a}|\V{y}^{1:K}) = \prod_{t}     \T{N}(\V{\mu}_{a}, \V{\sigma}_{a}^{2}), \label{eq:za}\quad 
 [\V{\mu}_{a}, \V{\sigma}_{a}^{2}] =
 \texttt{FC}([\V{a}_{i,s}, \gamma\V{a}_{i,d}^t]), \quad
 \gamma \sim \text{Bernoulli}(p) \label{eq:appearance}
 \end{eqnarray}
The mixing strategy helps to mitigate covariate shift and enforces the static component to learn the inherent semantics of the objects across frames. 


\begin{figure}[ht]
    \centering
\includegraphics[trim=40 90 30 20,clip,width=0.9\textwidth]{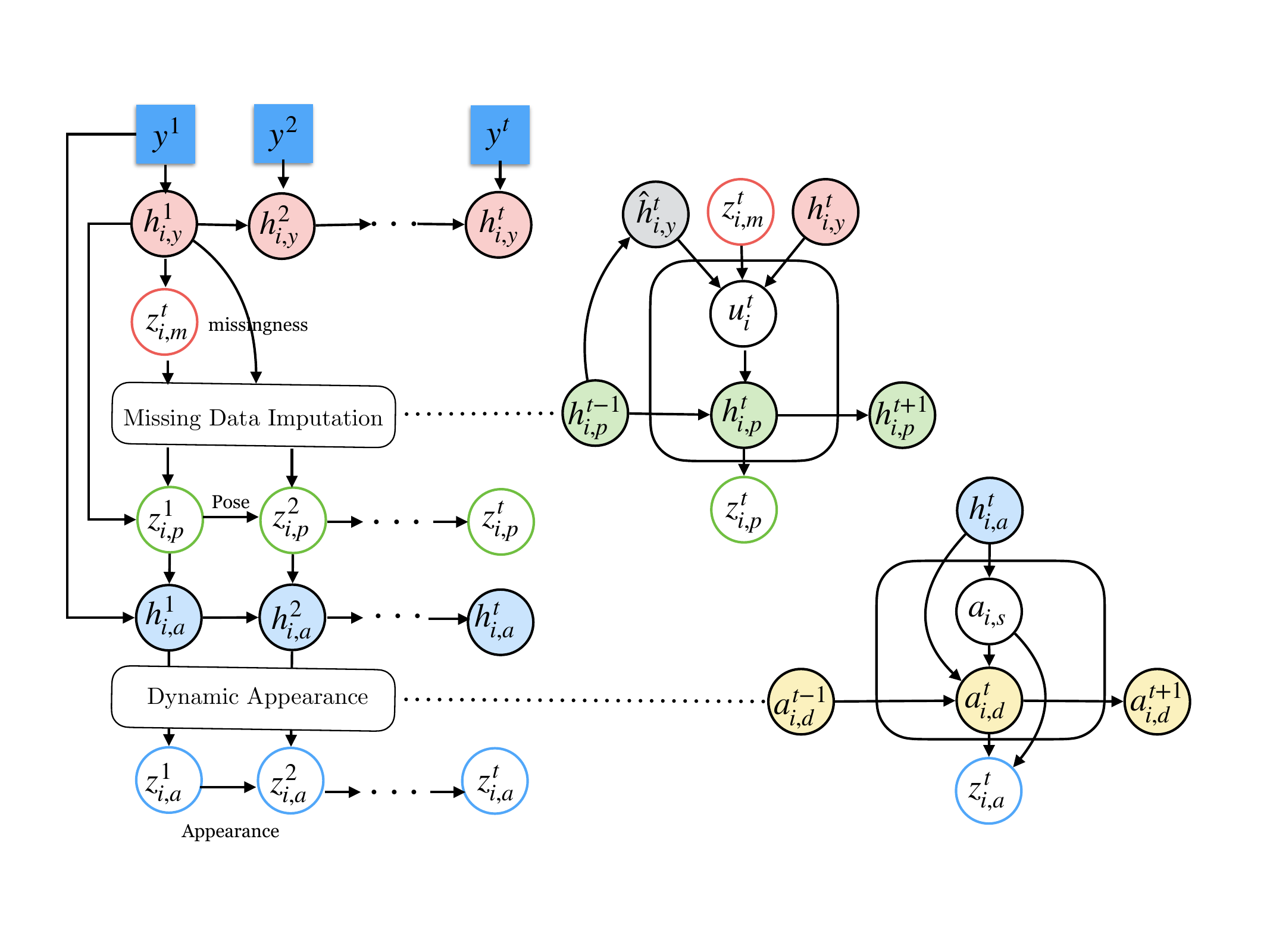}
    \caption{A graphical representation of \ours{}. From top to bottom:  inference of the missingness variable  $\V{z}_{i,m}^t$, missing data imputations model, inference of  the pose vectors  $\V{z}_{i,p}^t$ and  appearance variable $\V{z}_{i,a}^t$ using dynamic appearance inference. 
    }\label{fig:arch_enc}
    \vspace{-3mm}
\end{figure}

\subsection{Generative Model and Learning }\label{ssec:gen}




Given a  video with missing data $(\V{y}^1,\cdots, \V{y}^t)$, denote   the  underlying complete video  as $(\V{x}^1, \cdots \V{x}^t)$. Then, the generative  distribution of the video sequence is given by:
%
\begin{eqnarray}
    p(\V{y}^{1:K},\V{x}^{K+1:T}|\V{z}^{1:T})= \prod_{i=1}^N p(\V{y}_i^{1:K}|\V{z}_i^{1:K})p( \V{x}_i^{K+1:T}|\V{z}_i^{K+1:T})
    \label{eq:gen}
\end{eqnarray}
%
In unsupervised learning of video representations,  we  simultaneously  reconstruct the input video and predict  future  frames.  Given the inferred latent variables, we generate  $\V{y}_i^t$ and predict $\V{x}_i^t$  for each  object sequentially. In particular, we first generate the rectified object in the center, given the appearance $\V{z}_{i,a}^t$. The decoder  is parameterized by a deconvolutional layer. After that, we apply an spatial transformer $\T{T}^{}$ to rescale and place the object according to the pose $\V{z}_{i,p}^t$. For each object, the generative model is:
\begin{eqnarray}
p(\V{y}_i^t|\V{z}_{i,a}^t) = \T{T}(f_{\text{dec}}(\V{z}_{i,a}^t); \V{z}_{i,p}^t)\circ  (1-\V{z}_{i,m}^t), \quad
p(\V{x}_i^t|\V{z}_{i,a}^t) = \T{T}(f_{\text{dec}}(\V{z}_{i,a}^t),  \V{z}_{i,p}^t)
\label{eq:gen_model}
\end{eqnarray}
Future prediction is similar to  reconstruction, except we assume the video is always complete. The generated frame $\V{y}^t$ is the summation over $\V{y}_i^t$ for all objects.
Following the VAE framework, we train the model by maximizing the evidence lower bound (ELBO). Please see details in Appendix \ref{sec:app-ELBO} .

\section{Experiments}


\subsection{Experimental Setup}
We evaluate our method on variations of moving MNIST and MOTSChallenge multi-object tracking datasets. The prediction task is to generate 10 future frames, given an input of 10 frames. 
The baselines include the established state-of-the-art video prediction methods based on disentangled representation learning: 
\textsc{drnet} \cite{denton2017unsupervised}, \textsc{ddpae} \cite{ddpae} and \textsc{sqair} \cite{babaeizadeh2018stochastic}. 
\paragraph{Evaluation Metrics.} We use common evaluation metrics for  video quality on the visible pixels, which include pixel-level Binary Cross entropy (BCE) per frame, Mean Square Error (MSE) per frame, Peak Signal to Noise Ratio (PSNR) and Structural Similarity (SSIM). Additionally, \ours{} is a probabilistic model, hence we also report Negative Evidence Lower Bound (NELBO).

As our \textsc{dive} model simultaneously imputes missing data and generates improved predictions,  we report reconstruction and prediction performances separately.  For implementation details for the experiments, please see Appendix \ref{app:model_details}.

\subsection{Moving MNIST Experiments} \label{ssec:mnist-exp}
\paragraph{Data Description.}
Moving MNIST \cite{srivastava2015unsupervised} is a synthetic dataset consisting of two digits with size $28 \times 28$ moving independently in a $64 \times 64$ frame. Each sequence is generated on-the-fly by sampling MNIST digits and synthesizing trajectories with fixed velocity with randomly sampled angle and initial position. We train the model for 300 epochs in scenarios 1 and 2, and 600 epochs in scenario 3. For each epoch we generate 10k sequences. The test set contains 1,024 fixed sequences. We simulate a variety of missing data scenarios including:

\begin{itemize}[leftmargin=*,nolistsep]
  \item \textit{Partial Occlusion}: 
  we occlude the upper $32$ rows of the $64 \times 64$ pixel frame to simulate  the effect of objects being partially occluded at the boundaries of the frame. 
    \item \textit{Out of Scene}: 
    we randomly select an initial time step  $t’ = [3,9]$  and remove the object from the frame in steps $t’$ and $t’+1$ to simulate the out of scene phenomena for two consecutive steps.
    \item \textit{Missing with Varying Appearance}: we apply an elastic transformation \cite{1227801} to change the appearance of the objects individually. The transformation grid is chosen randomly for each sequence, and the parameter $\alpha$ of the deformation filter is set to $\alpha = 100$ and reduced linearly to 0 (no transformation) along the steps of the sequence. We remove each object for one time-step following the same logic as in scenario 2.
\end{itemize}

\begin{figure}[!t]
\begin{center}
\includegraphics[trim=30 10 0 0,clip,width=\textwidth]{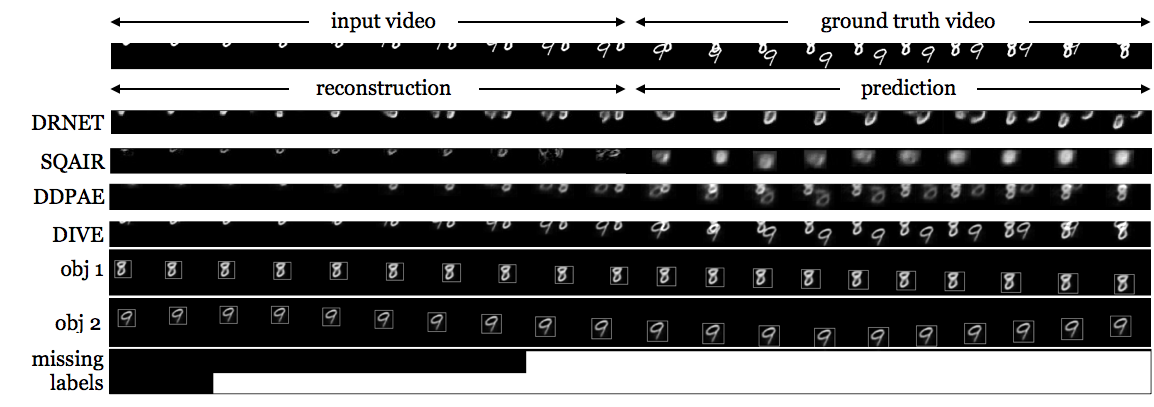}
\caption{\ac{change to better example} Partial missing qualitative results. \textit{Obj 1} and \textit{Obj 2} show \ours{}s individual object generations and \textit{missing labels} indicate whether each object is estimated completely missing in the scene. Note that objects are well decomposed, sharply generated and the labels properly predicted. }\label{fig:partial-exp}
\end{center}
\vspace{-3mm}
\end{figure}

\vspace{-3mm}
\paragraph{Scenario 1: Partial  occlusion. }
The top portion of Table \ref{table:exp} shows the quantitative performance comparison for all methods for the partial occlusion scenario. 
Our model outperforms all baseline models, except for the BCE in prediction. 
This is because \textsc{dive} generates sharper shapes which, in case of misalignment with the ground truth, have a larger effect on the pixel-level BCE. For reconstruction, our method often outperforms the baselines by a large margin, which highlights the significance of missing data imputation. Note that \textsc{sqair} performs well in reconstruction but fails in prediction. Prolonged full occlusions cause \textsc{sqair} to lose track of the object and re-identifying it as a new one when it reappears. 
%
Figure \ref{fig:partial-exp} shows a visualization of the predictions from \textsc{dive} and the baseline models. The bottom three rows show the decomposed representations from \ours{} for each object and the missingness labels for objects in the corresponding order. We observe that \textsc{drnet} and \textsc{sqair} fail to predict the objects position in the frame and appearance while \textsc{ddpae} generates blurry predictions with the correct pose. These failure cases rarely occur for \textsc{dive}. 
%
\begin{table}[b!]
  \caption{Quantities comparison of all methods for three missing scenarios w.r.t. reconstruction and prediction. From top to bottom: partial occlusion, out of scene and complete missing with varying appearance. The improvements of our method DIVE are evident for all scenarios.}
  \label{table:exp}
  \centering
  \resizebox{0.95\textwidth}{!}{ 
  \begin{tabular}{llllllllll}
    \toprule
        
        Scenario 1& \multicolumn{2}{c}{BCE $\downarrow$} 
            & \multicolumn{2}{c}{MSE $\downarrow$} 
                & \multicolumn{2}{c}{PSNR $\uparrow$} 
                    & \multicolumn{2}{c}{SSIM $\uparrow$}   
                        & NELBO $\downarrow$
                            \\
    \cmidrule(r){2-3}\cmidrule(r){4-5}\cmidrule(r){6-7}\cmidrule(r){8-9}
   Model &Rec&Pred&Rec&Pred&Rec&Pred&Rec&Pred&  \\
    \midrule
    \textsc{drnet}\cite{denton2017unsupervised}    & 482.07 & 852.59 & 72.21  & 96.36 & 7.99&6.89&0.76&0.72& / \\
    \textsc{sqair}\cite{kosiorek2018sequential}   &   178.71 & 967.20    & 21.84& 84.73 &13.19&9.96&\textbf{0.90}&0.73& -0.16 \\
    \textsc{ddpae}\cite{ddpae}   &   182.66 & \textbf{417.00}    & 39.09& 67.41 &17.56&15.49&0.77&0.72& -0.09 \\
    \textbf{\textsc{dive}}  &   \textbf{119.25}     &  459.10 &\textbf{19.73}& \textbf{64.49}  & \textbf{20.64}&\textbf{15.85}&\textbf{0.90}&\textbf{0.78}& \textbf{-0.18}\\
    \toprule
     Scenario 2\\
      \midrule
    \textsc{drnet}   & 392.33 & 1402.45 & 90.64  & 187.72 &9.59&9.88&0.80&0.67& / \\
     \textsc{sqair}  &   468.22 & 927.09    & 73.13& 137.04 &10.33&8.21&0.84&0.69&  -0.17 \\
    \textsc{ddpae}   &   266.03 &409.26     & 58.37& 89.57 &18.64&16.94&0.87&0.77& -0.17 \\
    \textbf{\textsc{dive}}  &   \textbf{165.42} &  \textbf{321.29}  &\textbf{27.03}& \textbf{64.17}  & \textbf{22.15}
                    &\textbf{18.56}&\textbf{0.93}&\textbf{0.83}& \textbf{-0.21}\\
    \toprule
    Scenario 3\\
    \midrule
    \textsc{drnet}   & 421.72 &1304.53 & 90.46  & 176.28 &9.91&7.33&0.75&0.70&  / \\
    \textsc{sqair}  &   560.51 & 1518.61    & 74.30& 163.25 &10.80&7.64&0.83&0.62&  -0.16 \\
    \textsc{ddpae}  &   322.23 & 403.48   & 63.63& 82.71 &18.29 &17.22&0.81&\textbf{0.78}& -0.18 \\
    \textbf{\textsc{dive}}  &   \textbf{272.74}     &  \textbf{374.59} &\textbf{42.81}& \textbf{74.87} 
                    & \textbf{20.08} &\textbf{17.61}&\textbf{0.87}& \textbf{0.78}&\textbf{-0.19}\\
    \bottomrule
  \end{tabular}
  }
\end{table}


\begin{figure}[t!]
\begin{center}
\includegraphics[trim=30 10 0 0,clip,width=\textwidth]{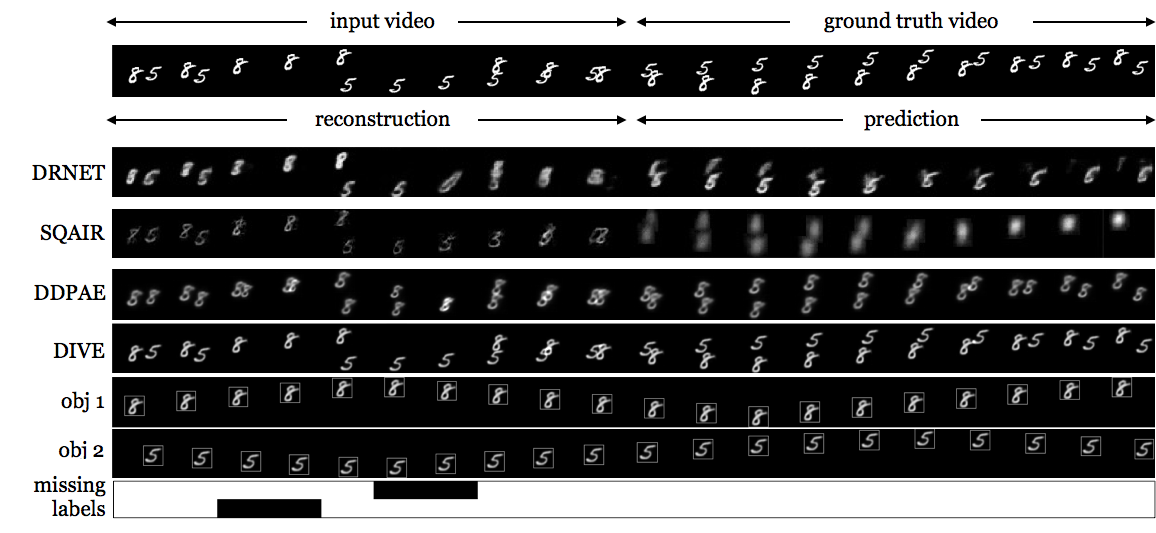}
\caption{Qualitative results for out of scene  missing scenario for two time steps.}\label{fig:comp-full-occ}
\end{center}
\vspace{-3mm}
\end{figure}

\begin{figure}[t!]
\begin{center}
\includegraphics[trim=30 10 0 0,clip,width=\textwidth]{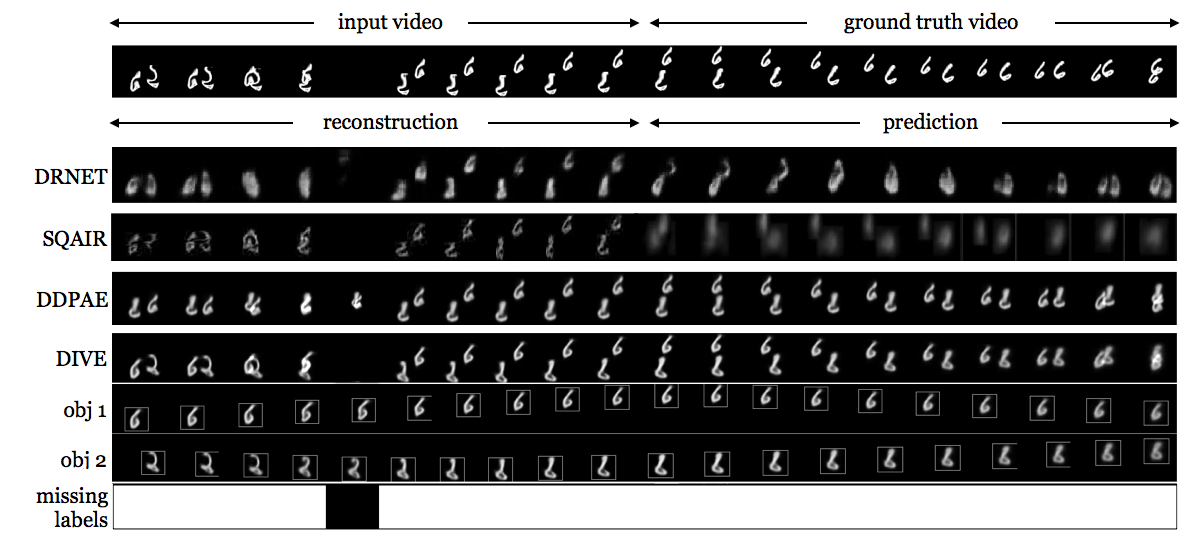}
\caption{Qualitative results for  one time step complete missing  with varying appearance.}\label{fig:varying-app}
\end{center}
\vspace{-3mm}
\end{figure}



\vspace{-3mm}
\paragraph{Scenario 2: Out of Scene.}
The middle portion of Table \ref{table:exp} illustrates the quantitative performance of all methods for scenario 2. 
We observe that our method achieves significant improvement across all metrics. 
This implies that our imputation of missing data is accurate and can drastically improve the predictions. 
Figure \ref{fig:comp-full-occ} shows the prediction results of all methods evaluated for the out of scene case. We observe that \textsc{drnet} and \textsc{sqair} fail to predict the  future pose, and the quality of the generated object appearance is poor.
The qualitative comparison with \textsc{ddpae} reveals that the objects generated by our model have higher brightness and sharpness. As the baselines cannot infer the object missingness, they may misidentify the missing object as any other object that is present. This would lead to confusion for modeling the pose and appearance. The figure also reveals how \ours{} is able to predict the missing labels and hallucinate the pose of the objects when missing, allowing for accurate predictions. 


\vspace{-3mm}
\paragraph{Scenario 3:  Missing with Varying Appearance.} 
Quantitative results for 1 time step complete missingness with varying appearance are shown in the bottom portion of Table \ref{table:exp}. 
Our  method again achieves the  best performance for all  metrics.
The difference between our models and baselines is quite significant given the difficulty of the task. Besides the complete missing frame, the varying appearances of the objects introduce an additional layer of complexity which can misguide the inference. 
Despite these challenges, \textsc{dive} can learn the appearance variation and successfully recognize the correct object in most cases. 
Figure \ref{fig:varying-app} visualizes the model predictions, a tough case where two seemingly different digits (``2'' and ``6'') are progressively transformed into the same digit (``6''). \textsc{sqair} and \textsc{drnet} have the ability to model varying appearance, but fail to generate reasonable predictions due to similar reasons as before. \textsc{ddpae} correctly predicts the pose after the missing step, but misidentifies the objects appearance before that. Also, \textsc{ddpae} simply cannot model appearance variation. \ours{} correctly estimates the pose and appearance variation of each object, while maintaining their identity throughout the sequence. 




\subsection{Pedestrian Experiments}
\begin{figure}[t!]
\begin{center}
\includegraphics[trim=40 10 0 0,clip,width=\textwidth]{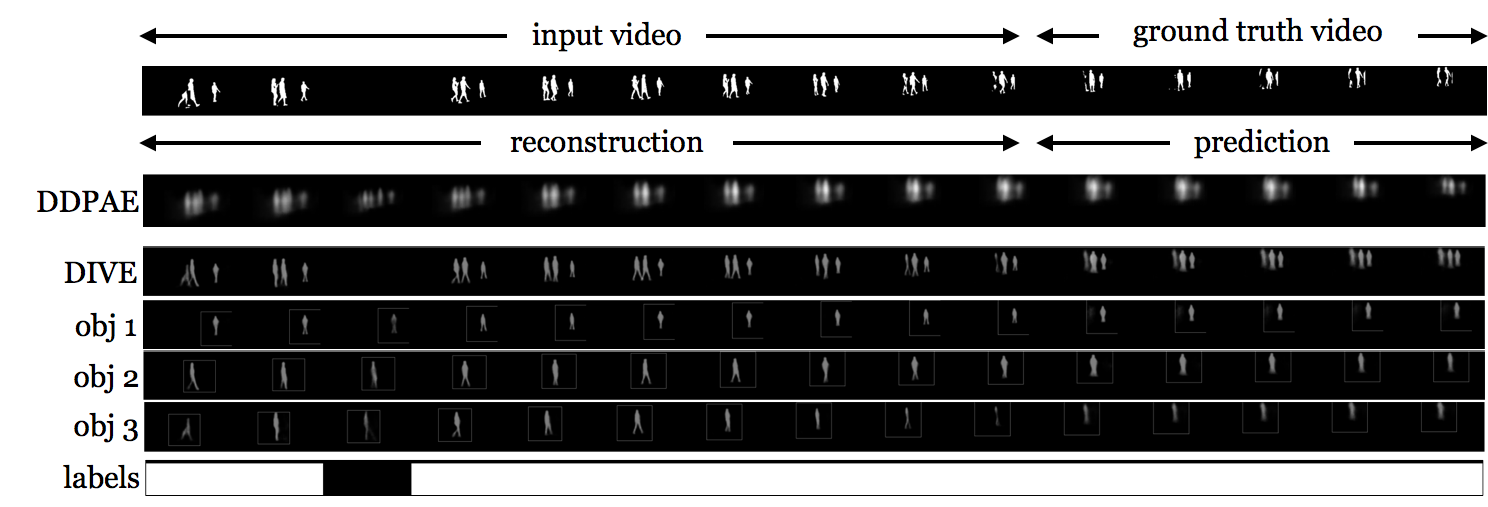}
\caption{MOTS data set qualitative results. Note that our method successfully identifies the missing time step, decomposes the objects and keeps track of the missing pedestrians.}\label{fig:ped-exp}
\end{center}
\vspace{-5mm}
\end{figure}
The Multi-Object Tracking and Segmentation (MOTS) Challenge \cite{MOT16} dataset consists of real world video sequences of pedestrians and cars. We use 2 ground truth sequences in which pedestrians have been fully segmented and annotated \cite{Voigtlaender19CVPR_MOTS}.  The annotated sequences are further processed into shorter 20 frame sub-sequences, binarized and with at most 3 unique pedestrians. The smallest objects are scaled and the sequences are augmented by simulating constant camera motion and 1 time step complete camera occlusion, further details deferred to Appendix \ref{app:exp_details}.

\begin{wraptable}{R}{0.55\textwidth}
  \centering
\vspace{-2mm}
     \caption{ Quantitative comparison on MOTS pedestrian dataset for \textsc{ddpae} and \textsc{dive}.}
  \label{table:ped_table}
 \resizebox{0.55\textwidth}{!}{ 
 \begin{tabular}{llllll}
     \toprule
     Model
     & BCE $\downarrow$
            & MSE $\downarrow$
                    & PSNR $\uparrow$
                            & SSIM $\uparrow$
                            & NELBO $\downarrow$
                            \\
    \midrule
    \textsc{ddpae}   &   2495.08 & 560.37   & 22.22& 0.90 &-0.24 \\
    \textsc{dive}  &   \textbf{1355.89}     &  \textbf{328.96} &\textbf{24.82}& \textbf{0.96} 
                    & \textbf{-0.26} \\
    \bottomrule
  \end{tabular}
  }
  \vspace{-3mm}
\end{wraptable}
Table \ref{table:ped_table} shows the quantitative metrics compared with the best performing baseline \textsc{ddpae}. This dataset mimics the missing scenarios 1 (partial occlusion) and 3 (missing with varying appearance) because the appearance walking pedestrians is constantly changing.  \textsc{dive} outperforms \textsc{ddpae} across all evaluation metrics. Figure \ref{fig:ped-exp} shows the outputs from both models as well as the decomposed objects and missingness labels from \textsc{dive}. Our method can accurately recognize 3 objects (pedestrians), infer their missingness and estimate their varying appearance. \textsc{ddpae} fails to decompose them due to its rigid assumption of fixed appearances and the inherent complexity of the scenario. 
In Appendix \ref{app:ablation}, we perform two ablation studies. One on the significance of dynamic appearance modeling, and  the other on the importance of estimating missingness and performing imputation.



\section{Conclusion and Discussion}
We propose a novel deep generative model  that can  simultaneously  perform object decomposition, latent space disentangling, missing data imputation, and video forecasting. The key novelty of our method includes missing data detection and imputation in the hidden representations, as well as a robust way of dealing with dynamic appearances.
Extensive experiments on moving MNIST demonstrate that \ours{} can impute missing data without supervision and generate videos of significantly higher quality. 
Future work will focus on improving our model so that it is able to handle the complexity and dynamics in  real world videos with unknown object number and colored scenes.


\section*{Broader Impact}

Videos provide a window into the physics of
the world we live in. They contain abundant visual information of what  objects are, how they move, and what happens when cameras move against the scene. Being able to learn a representation that disentangles these factors is fundamental to AI that can understand and act in spatiotemporal environment.
Despite the wealth of methods for  video prediction, state-of-the-art approaches are sensitive to missing data, which are very common in real-world videos. Our proposed model significantly improves the robustness of video prediction methods against missing data, and thereby increasing the practical values of video prediction techniques and our trust in AI. %
Video surveillance systems can be potentially abused for discriminatory targeting, and we remained cognizant of the bias in our training data. To reduce the potential risk of this, we pre-processed the MOTSChallenge videos to greyscale.

\begin{ack}
This work was supported in part by  NSF under Grants IIS\#1850349, IIS\#1814631, ECCS\#1808381 and CMMI\#1638234, the U. S. Army Research Office under Grant W911NF-20-1-0334 and the Alert DHS Center of Excellence under Award Number 2013-ST-061-ED0001. The views and conclusions contained in this document are those of the authors and should not be interpreted as necessarily representing the official policies, either expressed or implied, of the U.S. Department of Homeland Security. We thank Dr. Adam Kosiorek for helpful discussions. Additional revenues related to this work:  ONR \# N68335-19-C-0310, Google Faculty Research Award, Adobe Data Science Research Awards,   GPUs donated by NVIDIA, and computing allocation awarded by DOE.
\end{ack}

\bibliographystyle{unsrt}
\bibliography{neurips_2020.bib}

\clearpage

\normalsize
\appendix


\section{Model Implementation Details}
\label{app:model_details}

\paragraph{\textsc{sqair}} The \textsc{sqair} model is sensitive to hyper-parameters \cite{kosiorek2018sequential}. Different combinations of hyper-parameters are used to reproduce the best performance of the model. However, through our communication with the  authors, \textsc{sqair} model is not designed for the missing data scenario. Thus, we were not able to reach similar level of performance reported in \cite{kosiorek2018sequential}.  In order to obtain the best performance of \textsc{sqair} for our data set, we trained and evaluated the reconstruction model and prediction model separately as we found  that \textsc{sqair} model is more stable with the reconstruction task and thus could be trained longer (300 epochs and more). However, for the prediction task, the main issue we encountered was that the gradients would vanish for some combinations of hyper-parameters  and the model was not able to make predictions after certain number of training epochs (this number can fluctuate). In order to obtain the best performance of \textsc{sqair}, we kept the model training  until it could  generate predictions and select the checkpoint with the best performance. We used rmsprop to optimize the \textsc{sqair} model and the model uses important-weighted auto-encoder\cite{Burda2016ImportanceWA} with 5 particles as a general structure. For more implementation details, please refer to the github of \textsc{sqair}\cite{kosiorek2018sequential}. It is also important to note that the training time per 100 epochs for \textsc{sqair} is at least 5 times more than the training time of our DIVE model.

\paragraph{\textsc{drnet}} The original version of \textsc{drnet} model only uses the first four frames for training. In order to adapt \textsc{drnet} for our prediction task, we changed the scene discriminator in \textsc{drnet} to train on all frames in the sequences. This modification is more suitable for our missing data scenario. Because if there were missing data in the first four frames, the scene discriminator trained only on the first four frames would easily fail. However, after this modification, the probability for the scene discriminator to successfully recognize the scene also increases. Except for this modification, the rest of the model was kept exactly the same as the author's implementation for better reproduction of results. It is also important to note that the main network and the \textsc{lstm} in \textsc{drnet} were trained separately.  The main network was trained first and then the \textsc{lstm} was trained based on results of the main network. Therefore, if the main network failed to recognize the objects, the \textsc{lstm} would also fail to learn the trajectories. We used the default Adam optimizer in \textsc{drnet} to train the model. The scene discriminator was trained with BCE loss. The main network and \textsc{lstm} were trained with MSE loss. For more implementation details, please refer to the github of \textsc{drnet}\cite{denton2017unsupervised}.

\paragraph{\textsc{ddpae}} We used the code provided by the authors. The hyperparameters that they use in the public version were kept unchanged. Also we followed the instructions in their github repository (\url{https://github.com/jthsieh/DDPAE-video-prediction}) for the Moving MNIST experiment. However, for some of the experiments, we have added additional features  to produce better results. 
For the Pedestrian experiments, we aligned the hyperparameters that are semantically similar with \ours{} implementation. Also, the pose size was constrained less than in the default setting. This way, the model can adapt to a highly varying dataset.

\paragraph{\textsc{dive}} The main variables have the following dimensions: $\V{z}_{i,a}^t \in \mathbb{R}^{128}$, $\V{a}_{i,s} \in \mathbb{R}^{256}$, $\V{a}_{i,d}^t \in \mathbb{R}^{48}$, $\V{z}_{i,p}^t \in \mathbb{R}^3$, $\V{z}_{i,m}^t \in \mathbb{Z}^1$ and $\V{h}_{i,y}^t \in \mathbb{R}^{64}$. The dimensions were chosen after a manual sweep of hyperparameters range. Particularly, the dimensionality of $\V{a}_{i,d}^t$ was chosen from the range $[12, 24, 48, 64]$;  $\V{z}_{i,a}^t$ and $\V{a}_{i,s}$ from $[64, 128, 256]$; and $\V{h}_{i,y}^t$ from $[48,64,96]$. The learning rate was set to $10^{-3}$ and reduced to $4^{-3}$ at $1/3$ of the training iterations, and we used a batch size of $64$. The Bernoulli distribution for the imputation model has probability $p=0.25$ in training and $p=0$ in testing. For the appearance model, the Bernoulli distribution has $p=0.7$, which was increased to $p=0.85$ after 3,000 iterations during training. For testing, we set $p=1$. In both cases, we experimented with different values of p ($\pm 0.1$) and did not find any significant difference.
The number of objects $N$ is specified a priori, but only as an upper bound. For the MOTS dataset, we set $N=3$ but the actual number of objects is often lower. For all the Moving MNIST experiments we set $N=2$. Similarly to DDPAE, our model can learn to set the redundant components to be empty. Further details can be found in the provided codebase.

The missingness latent variable, $\V{z}_{i,m}^t$, was implemented with a Heaviside step function in the pose encoding model, with a $-0.5$ bias in the logit. However, to allow the  gradients to propagate, the Heavyside function $H(x)$  is not a variable node in our computational graph, hence we are not differentiating through it.  We use \textit{torch.where()} function in Pytorch to implement this condition operation.
As shown in Figure \ref{fig:arch_all} (top) and Equation \ref{eq:gen_model}, $\V{z}_{i,m}$ is a masking indicator. In practice, we multiply each decoded object by the logit before the Heavyside function instead of the binary label. Hence, $\V{z}_{i,m}$ gets its gradients from the decoder.

To adapt to three missing scenarios, we made minor changes to our implementation. 
For missing scenario 1 (partial occlusion) and 2 (out of scene) of the MovingMNIST experiments, because the objects appearance remain static, we did not include the dynamic appearance model component. The appearance encoding is therefore adjusted accordingly.  We followed Equation \ref{eq:stat-app} to generate the static appearance, but we skipped the input frames $\V{y}^t$ and hidden states $\V{h}^t_{i,a}$ in $\texttt{LSTM}_1$ where we predicted missingness $\V{z}_{i,m}^t=1$.
For  partial occlusion training with Moving MNIST dataset, we used a scheduling mechanism to evaluate the loss only for the visible area of the frame. We applied the same procedure to all the baselines for a fair comparison.
For the pedestrian dataset, similarly to \textsc{ddpae}, we relaxed the pose size constraint to accommodate the highly dynamic pose size in real-world videos.
With this implementation, we measure the training time. It takes 91 minutes to carry out 100 epochs, for which we process 1 million samples in batches of 64.

\paragraph{Software} We implemented this method using Ubuntu 18.04, Python 3.6, Pytorch 1.2.0, Cuda 10.0 and Pyro 0.2 as a framework for probabilistic programming.

\paragraph{Hardware} For each of our experiments we used 1 GPU RTX 2080 Ti (Blower Edition) with 12.8GB of memory.


\section{Datasets Details} \label{app:exp_details}

\paragraph{Moving MNIST with elastic deformation.} 
In order to simulate slowly varying appearance in Scenario 3, we applied an elastic deformation to the objects in the scene. Given a uniform grid that represents the object pixel coordinates, we generated a distortion. We created  a displacement random field with parameters $\alpha$ and $\sigma$. These parameters controlled the intensity of the deformation and the smoothing of the field, respectively. 
The displacement field was added to the uniform grid, and used to deform the coordinates of the given digit. This is described in \cite{1227801}. The transformation was done independently to every digit.
We set $\sigma=4$ and $\alpha$ varied linearly from $100$ to $0$ along the sequence.

\paragraph{MOTS Challenge pre-processing.} 
The Multi-Object Tracking and Segmentation (MOTS) Challenge \cite{MOT16} dataset focuses on the task of multi-object tracking to multi-object tracking and segmentation. It provides dense pixel-level annotations for two existing tracking datasets. It comprises 65,213 pixel masks for 977 distinct objects (cars and pedestrians) in 10,870 video frames. 
For our task, we used 2 scenes with only pedestrians. Each one of these scenes was processed as follows: We kept the dense annotations as the shapes of the objects, and discarded all remaining content (such as the background). Given the large variance in the objects size, we resized the objects below the average size in the scene to the average and added a small random margin. Each scene was divided in sequences of 20 frames, reducing the sampling rate by a factor of 5 to increase displacements of objects. For each sequence, we selected all combinations of 3 objects to augment the data. We binarized the grey values of all sequences.
Each sequence was padded randomly to fit a square and resized to $256 \times 256$ pixels. Finally, during training we added on-the-fly transformation to the clips. We subtracted all content for one random time step and sequentially affine-transformed the frames  to simulate full camera occlusion and constant camera motion. This was also done when generating the fixed testing sequences.
As a result, we used 4,416 sequences for training and 675 for testing, while making sure they belonged to different scenes.

\section{Ablation Study} \label{app:ablation}

\subsection{Appearance variability}
\begin{figure}[htbp]
\begin{center}
\includegraphics[trim=0 0 0 0,clip,width=\textwidth]{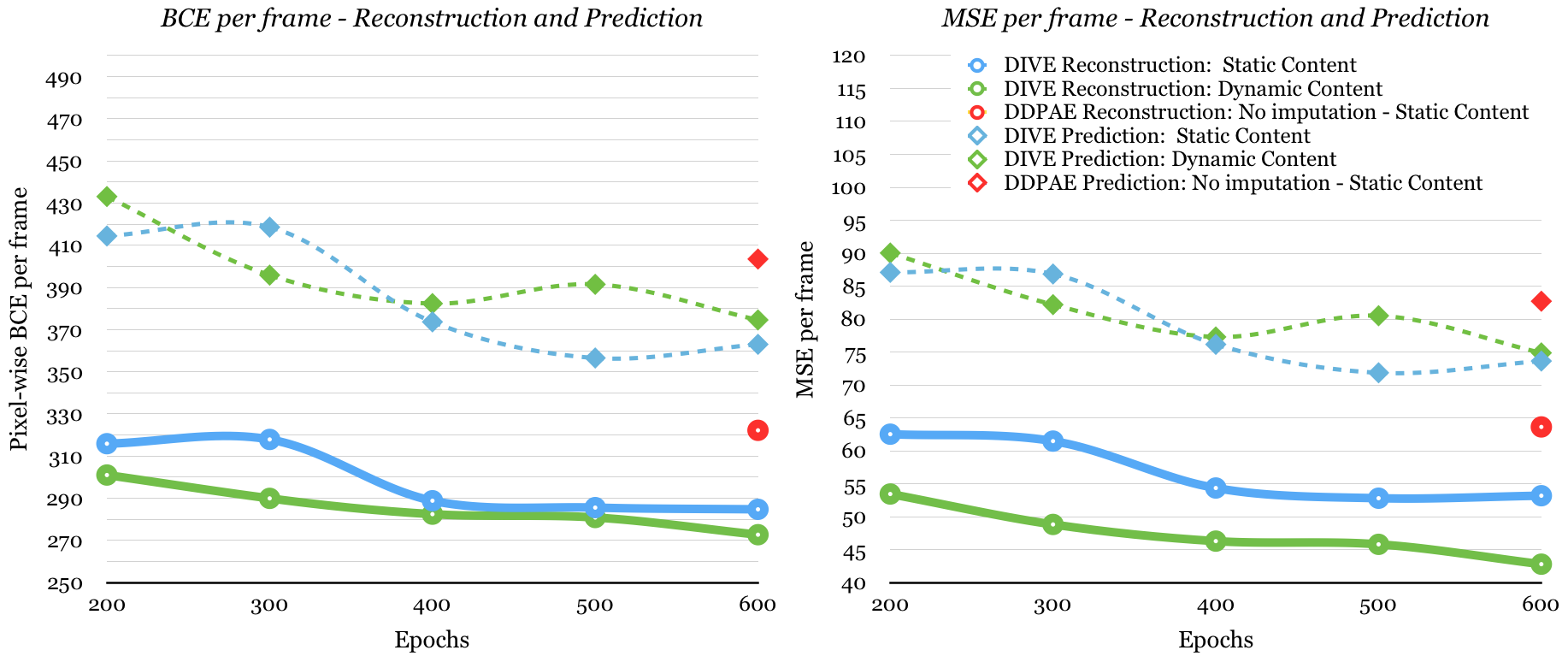}
\caption{Ablation study for static and dynamic appearance modeling for missing Scenario 3 of the Moving MNIST experiments. \textsc{ddpae} results were also shown for comparison purposes.}\label{fig:ablation}
\end{center}
\vspace{-3mm}
\end{figure}

In order to highlight the significance of dynamic appearance modeling, we performed an ablative study for \ours{}, focusing on Scenario 3 in Section \ref{ssec:mnist-exp}.  In particular, we  compared two cases:
(1) \textbf{dynamic appearance.} This is our main configuration. Missingness was estimated with hard labels, binarized with a step function while encoding.
    The appearance was modeled as in Equation \eqref{eq:appearance}, where the Bernoulli probability is $p=1$ in testing and therefore we explicitly modeled the dynamic appearance.
(2) \textbf{static appearance.} In this case, we altered the original configuration by setting $p=0$ for the Bernoulli distribution in Equation \ref{eq:appearance}. This allows only for static (constant) appearance generation.

For each case, we trained the model for 600 epochs and kept the model every $100$ epochs for the range $[200, 600]$, with the same training and testing setup as previously reported for this scenario. We used the \textsc{DDPAE} results trained for  600 epochs as a baseline. We tested the models and report BCE and MSE per frame metrics, separately for input reconstruction and output prediction.

Figure \ref{fig:ablation} shows the quantitative results for both BCE (left) and MSE (right). We can see that for reconstruction, having dynamic appearance components significantly reduces the reconstruction error, specially for MSE. This is because Scenario 3 contains digits with large distortion in the input. 
Hence more flexible appearance modeling better adapts  to the changing shapes.

However, 
predicting the sequence into the future inevitably introduces  uncertainty, leading to blurry predictions. Static  modeling captures the shared constant appearance, which have low intensity deformations. Therefore, it does not suffer large appearance variations, and generates sharper shapes  in prediction.   
%
%
However, as   the baseline \textsc{ddpae} does not provide a mechanism for missing data imputation or varying appearance, both of our approaches outperform  (\textsc{ddpae}) by a large margin, even in the early stages of training. 

We also conducted an ablative study on the missingness variable. In our implementation, we chose a heavy-sided function to obtain ``hard'' missingness labels. An alternative is 
 a \textit{Sigmoid} activation function to obtain ``soft'' label  $\V{z}_{i,m}^t \in (0,1)$. We tested both and found that the model can always  learn  the labels correctly. The performance difference was not statistically significant.


\subsection{Missingness Variable}

In our experiments, we consider \textsc{ddpae} to be the closest architecture to ours without missingness variables. To further validate the significance of explicitly modeling missingness, we also test the exact same \textsc{dive} architecture with and without the missingness variable, for Scenario 2 of Moving MNIST experiments. Table \ref{table:ablation-miss} shows the quantitative results after 300 epochs of training:

\begin{table}[h]
    \caption{Quantitative results for Scenario 2 of Moving MNIST experiments. We perform experiments with and without missing data label and imputation. Missing data imputation for \textsc{dive} shows clear improvements in all metrics for both reconstruction and prediction.} \label{table:ablation-miss}
    \centering
    \resizebox{.95\textwidth}{!}{ 
    \begin{tabular}{lllllllll}
        \toprule
            Mov. MNIST Scenario 2& \multicolumn{2}{c}{BCE $\downarrow$} 
            & \multicolumn{2}{c}{MSE $\downarrow$} 
                & \multicolumn{2}{c}{PSNR $\uparrow$} 
                    & \multicolumn{2}{c}{SSIM $\uparrow$}   
                            \\
        \cmidrule(r){2-3}\cmidrule(r){4-5}\cmidrule(r){6-7}\cmidrule(r){8-9}
        Model (trained 300 epochs) &Rec&Pred&Rec&Pred&Rec&Pred&Rec&Pred \\
        \midrule
        \textbf{\textsc{dive} w/o missingness}& 236.35&356.82&49.07&76.52&19.40&17.66&0.86&0.82 \\
        \textbf{\textsc{dive} w missingness}  &   \textbf{165.42} &  \textbf{321.29}  &\textbf{27.03}& \textbf{64.17}  & \textbf{22.15}&\textbf{18.56}&\textbf{0.93}&\textbf{0.83}\\
        \bottomrule
    \end{tabular}
    }
    \vspace{-0.5cm}
\end{table}

The results demonstrate   that explicitly reasoning about missingness and performing imputation is the key, not only to the  reconstruction but also to future predictions. This is partially attributed to a better learning of the underlying dynamics of the scene.

\section{Optimization Objective} \label{sec:app-ELBO}
The optimization objective is to maximize the evidence lower bound (ELBO), as in the common VAE framework:
\begin{eqnarray}
    \log p_{\theta}({\V{y}}^{1:K}, {\V{x}}^{K+1:T}) &\ge& \mathbb{E}_q\left[\log p_{\theta}\left({\V{y}}^{1:K}|{\V{z}}^{1:K}_{1:N}
    \right)-\text{KL}\left(q_{\phi}\left({\V{z}}^{1:K}_{1:N} \right) ||p\left({\V{z}}^{1:K}_{1:N}\right)\right) \right] 
   \\ 
    &+& \mathbb{E}_q\left[\log p_{\theta}\left({\V{x}}^{K+1:T}|\V{z}^{K+1:T}_{1:N} \right) -\text{KL}\left(q_{\phi}\left(\V{z}^{K+1:T}_{1:N} \right) || p\left(\V{z}^{K+1:T}_{1:N}\right)\right)\right]\nonumber
\end{eqnarray}
Here, \textsc{DIVE} uses self-supervision for reconstructing the corrupted input $\V{y}^{1:K}$ and predicting the complete output $\V{x}^{K+1:T}$. We add a regularization term to minimize the KL-divergence between our latent space representation and a Gaussian prior, parametrized by its mean and variance.
Note that $N$ is our prior on the number of objects in the scene.

\section{More examples and failure cases of DIVE}
In this section, we provide more examples including failure cases from three missing scenarios experiments. For each of the examples, the first 10 frames are the inputs, followed by the 10 predicted frames. The top row is the ground truth and the second to the last row is the reconstructions/predictions from \ours{}. We also show the decomposed objects and the learned missingness labels, respectively.

\begin{figure}[h]
\begin{center}

  \includegraphics[clip,width=\textwidth]{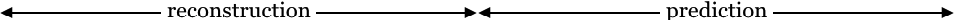}%
\vspace{-0.2cm}\\
    \subfloat[A failure case where our DIVE model cannot reconstruct and predict  digit ``7'', as it doesn't appear in the input.]{%
  \includegraphics[clip,width=\textwidth]{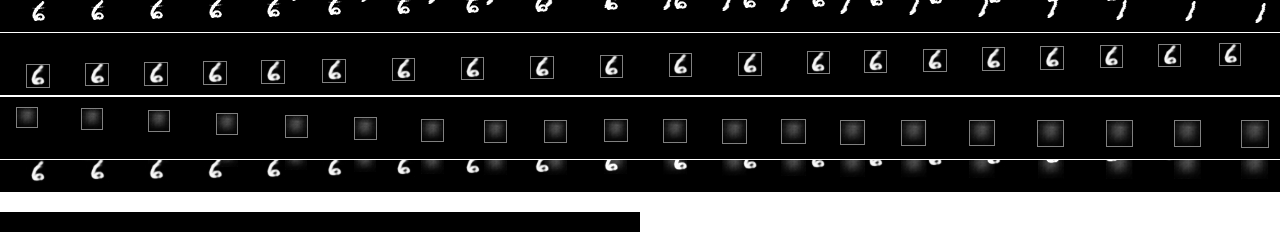}%
}\\
\subfloat[A success case where our model recovers the heavily corrupted digit.]{%
  \includegraphics[clip,width=\textwidth]{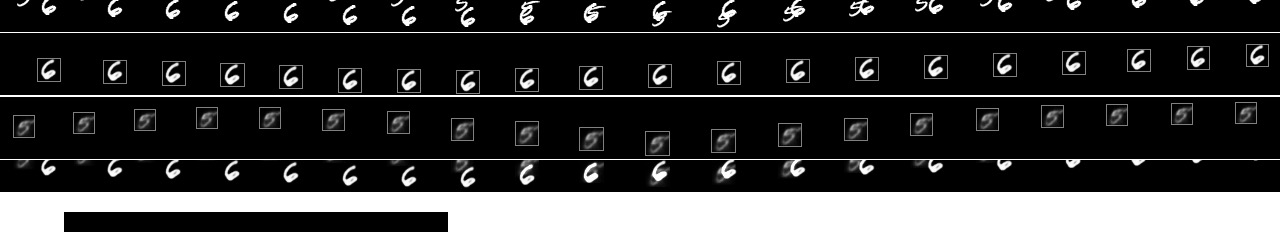}%
}\\

\end{center}
\caption{More examples for missing Scenario 1: Partial occlusion experiment. The rows for each figure from top to bottom are (1) ground truth, (2) first object,  (3) second object, (4) \ours{} predictions, (5) predicted missing labels for each object. We use the same display format for all Moving MNIST examples below.}\label{fig:app-partial}
\end{figure}

Figure \ref{fig:app-partial} shows three examples for missing scenario 1 (partial occlusion). Figure \ref{fig:app-partial}(a) shows a failure case where \ours{} cannot recognize and generate  digit ``7'' as it only reappears at the very end. 
This is partially due to our imputation mechanism, which only uses the previous information not the future information. 
%
%
Figure \ref{fig:app-partial}(b) shows a success case where even though one of the digits is heavily corrupted in the input frames, \ours{} could still reconstruct it in the results. In this case, digit ``5'' is totally missing in five input frames and is heavily corrupted or overlaps with the other digits in the rest of the input frames. Our DIVE model successfully reconstructs and predicts it in almost all frames. It is also important to note that the imputation of the missing digit five from second frame to seventh frame is smooth and accurate (see third row of the figure).

\begin{figure}[htbp]
\begin{center}
  \includegraphics[clip,width=\textwidth]{appendix_figure/arrows.png}%
\vspace{-0.2cm}\\
\subfloat[Failure case with two digits overlapping in all input frames.]{%
  \includegraphics[clip,width=\textwidth]{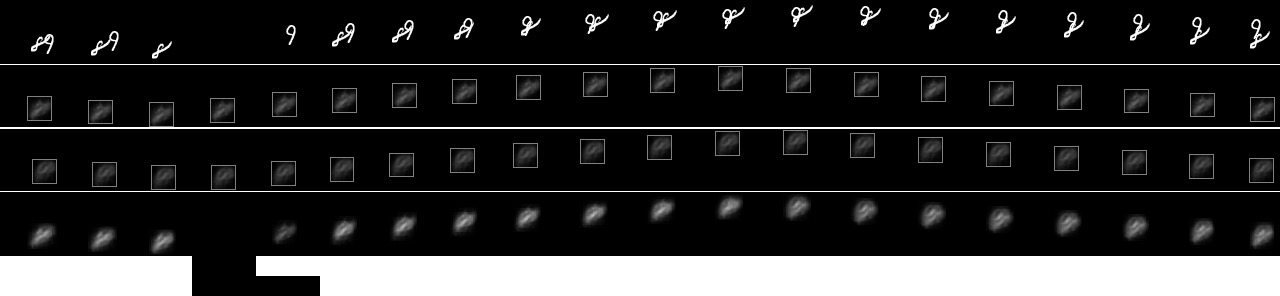}%
}\\

\subfloat[Success case with two digits overlapping frequently.]{%
  \includegraphics[clip,width=\textwidth]{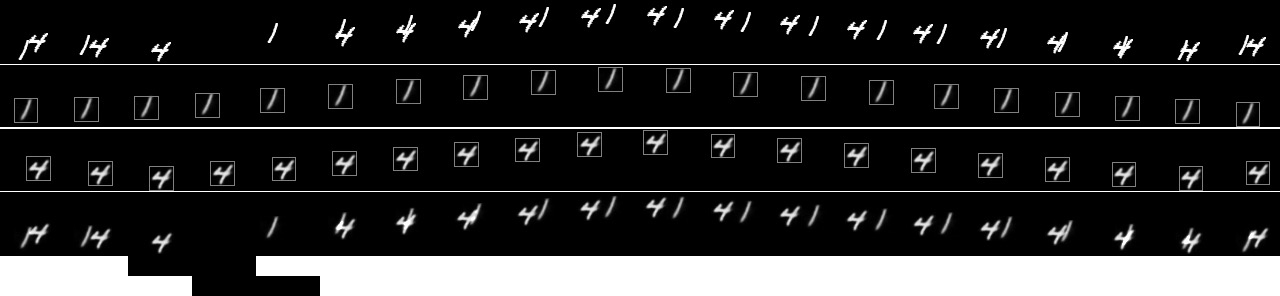}%
}\\
\caption{Examples for Scenario 2: Out of scene for two time steps.}\label{fig:app-complete}
\end{center}
\end{figure}

\begin{figure}[t!]
\begin{center}
  \includegraphics[clip,width=\textwidth]{appendix_figure/arrows.png}%
\vspace{-0.2cm}\\
\subfloat[A failure case where the model misrecognizes the digits.]{%
  \includegraphics[clip,width=\textwidth]{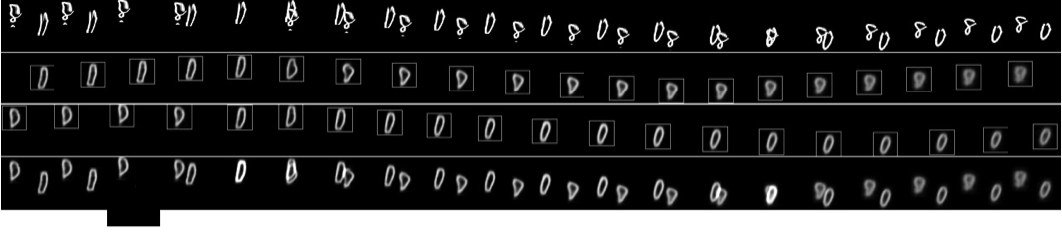}%
}\\
\subfloat[A success case on two similar digits.]{%
  \includegraphics[clip,width=\textwidth]{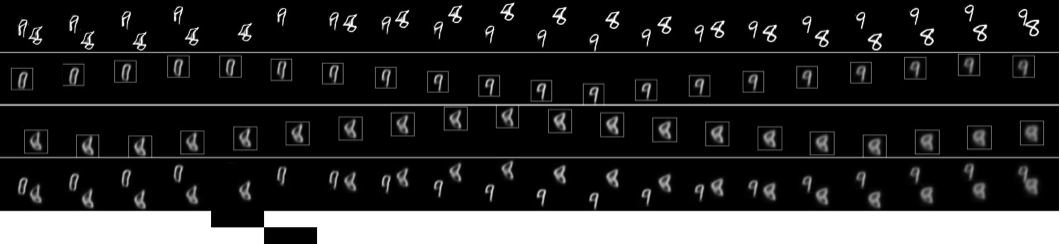}%
}\\
\caption{Examples for Scenario 3: varying appearance experiment. }\label{fig:app-vary}

\end{center}
\end{figure}

Figure \ref{fig:app-complete} shows more examples from missing Scenario 2 (out of scene). Specifically, a failure case where DIVE cannot recognize both of the digits is shown in Figure \ref{fig:app-complete}(a). In this case, the digits entangle with each other almost in every frame and thus the model recognizes them as one object. We also show a success case where the two digits entangle with each other frequently  in Figure \ref{fig:app-complete}(b). From these two cases,  we can conclude that the model needs as least one frame where the two digits are separable to generate decent results.

Figure \ref{fig:app-vary} displays more examples for an experiment on missing Scenario 3 (complete missing with varying appearance).  Figure \ref{fig:app-vary}(a) shows a failure case where after the fifth frame, our \ours{} model mis-recognizes the two digits. The switching happens when in the fifth frame, digit ``8'' is missing from the scene and the digit ``8'' and ``0'' have similar appearances. After the switching, the model fails to  recover the initial assignment of the objects. Although in this example our model generates decent results, we cannot overlook the potential issue. Especially when the trajectories of objects are very complex and heterogeneous, confusion in appearances  could lead to  inaccurate predictions of trajectories. Figure \ref{fig:app-vary}(b) shows a success case where the two digits are similar.

\begin{figure}[h]
\begin{center}
\subfloat[A failure case with a split object.]{%
  \includegraphics[clip,width=\textwidth]{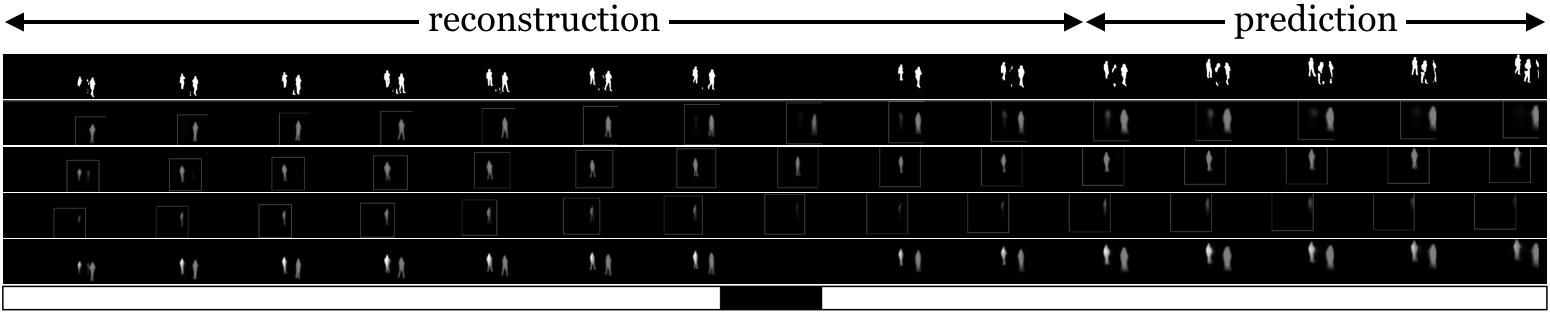}%
}\\
\subfloat[A success in a difficult case of overlapping objects.]{%
  \includegraphics[clip,width=\textwidth]{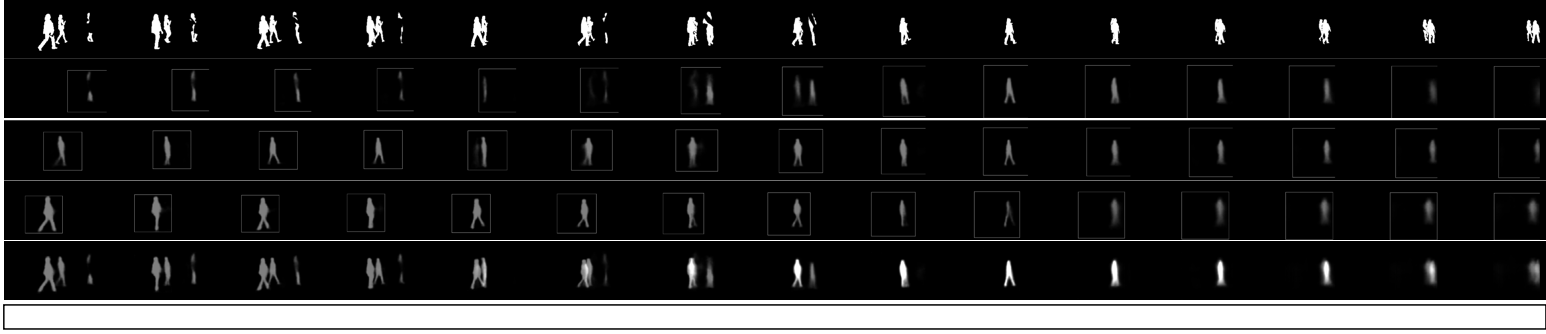}%
}\\
\caption{Qualitative examples for pedestrian (MOTS) dataset. The rows from top to bottom are: (1) ground truth, (2) first object, (3) second object, (4) third object, (5) \ours{} prediction, (6) combined predicted missing labels for each object.}
\label{fig:app-ped}
\end{center}
\end{figure}

More examples from the MOTS data set are shown in Figure \ref{fig:app-ped}. Figure \ref{fig:app-ped}(a) shows a failure example where the object/pedestrian is partly present in the $4_{th}$ row, that should be empty. Given the low displacement of the objects, the model sometimes has problems to infer which entities are independent. This can create duplicated content when we decompose the frames. It can also happen, that two objects that are static or move at the same velocity are encoded as a single entity. The failure case also shows how the model can't predict the appearance of a new object that hasn't been identified in the input. This is not surprising, as a human wouldn't have been able to make such prediction. Figure \ref{fig:app-ped}(b) shows a success case where our model encodes each pedestrian properly and generates reasonable predictions. This case is especially difficult because, although there is no full frame missing, two of the objects overlap for several frames at the input sequence.
\end{document}